\documentclass{article}

\usepackage{arxiv}      

\usepackage{array}

\usepackage{booktabs}
\usepackage[table]{xcolor}
\usepackage{makecell}
\usepackage{amsmath}
\usepackage{graphicx}
\usepackage{makecell}  
\usepackage{needspace}

\usepackage{caption}
\usepackage{wrapfig}     
\usepackage[utf8]{inputenc}

\usepackage{xparse}      
\usepackage{fancyvrb}    
\usepackage{fvextra}     

\NewDocumentCommand{\inputminted}{O{} m m}{%
  \VerbatimInput[
    fontsize=\small,
    frame=single,
    breaklines=true
  ]{#3}%
}

\newcommand{\setminted}[1]{}

\usepackage{listings}
\usepackage[utf8]{inputenc}
\usepackage[T1]{fontenc}
\usepackage{upquote}  

\usepackage{xcolor}

\definecolor{codebg}{HTML}{F7F7F8}
\definecolor{coderule}{HTML}{E5E7EB}
\definecolor{codekw}{HTML}{1F6FEB}
\definecolor{codestr}{HTML}{0EA5E9}
\definecolor{codecm}{HTML}{6B7280}
\definecolor{codenums}{HTML}{9CA3AF}

\usepackage[utf8]{inputenc} 
\usepackage[T1]{fontenc}    
\usepackage{url}            
\usepackage{booktabs}       
\usepackage{amsfonts}       
\usepackage{nicefrac}       
\usepackage{microtype}      
\usepackage{xcolor}         
\usepackage{multirow} 
\usepackage{adjustbox}

\definecolor{wacvblue}{rgb}{0.21,0.49,0.74}
\usepackage[colorlinks,breaklinks,pagebackref,linkcolor=magenta,citecolor=magenta,urlcolor=magenta]{hyperref}
\usepackage[capitalize]{cleveref}

\title{TRAVL: A Recipe for Making Video-Language Models Better Judges of Physics Implausibility}

\author{
  Saman Motamed$^{1,2}$ \quad
  Minghao Chen$^{2}$ \quad
  Luc Van Gool$^{1}$ \quad
  Iro Laina$^{2}$ \\
  $^1$ INSAIT, Sofia University "St. Kliment Ohridski", Bulgaria \\
  $^2$ Visual Geometry Group, University of Oxford \\ 
  \\
   \href{https://sam-motamed.github.io/projects/TRAVL}{sam-motamed.github.io/projects/TRAVL}\\[0.8em]
}

\begin{document}
\makeatletter
\twocolumn[{%
\begin{@twocolumnfalse}
\maketitle

\centering
\includegraphics[width=\textwidth,height=0.28\textheight,keepaspectratio]{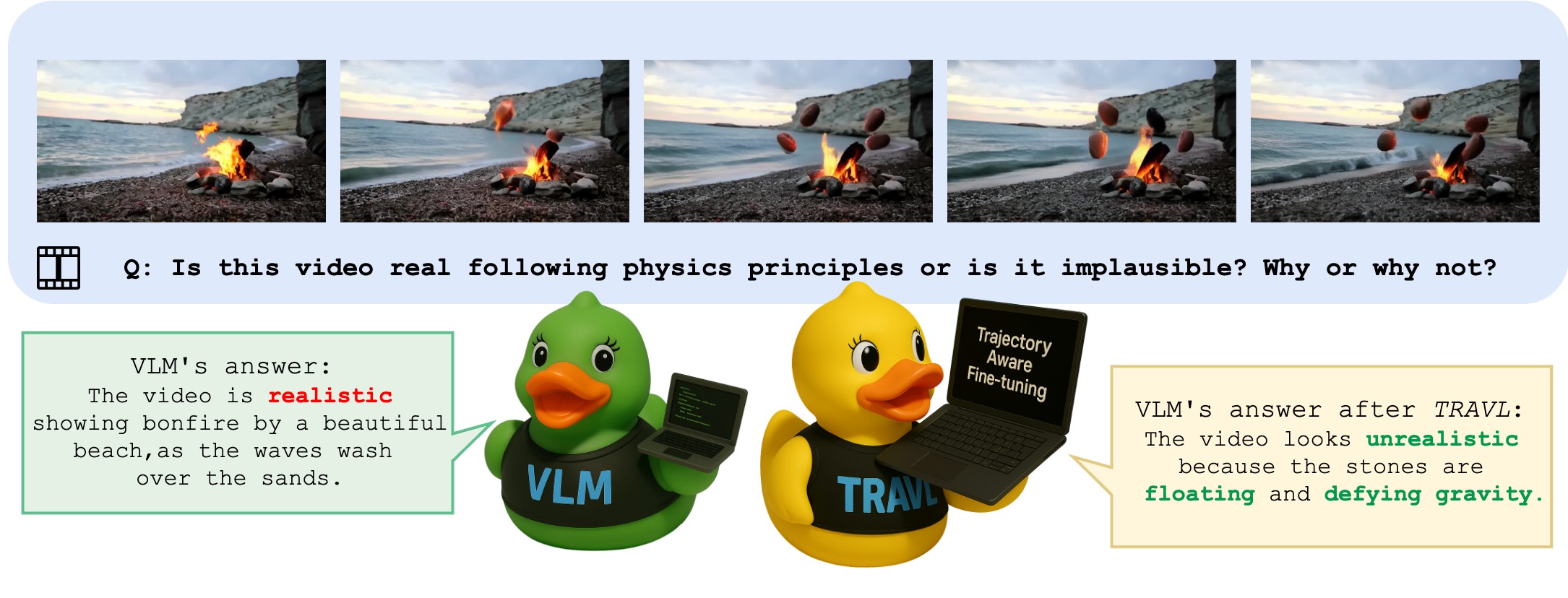}
\captionof{figure}{Video Language Models (VLMs) often struggle with fine-grained understanding of physics realism. We propose a fine-tuning recipe that helps VLMs become better judges of physics implausibility.}
\label{fig:teaser}

\begin{abstract}
Despite impressive visual fidelity, modern video generative models frequently produce sequences that violate intuitive physical laws, such as objects floating, teleporting, or morphing in ways that defy causality. While humans can easily detect such implausibilities, there remains no robust method for quantitatively assessing physical realism in video. In this work, we explore whether Video-Language Models (VLMs) can be trained to serve as reliable judges of physical plausibility. We find that existing VLMs struggle to identify physics violations, exposing fundamental limitations in their temporal and causal reasoning. To address this, we introduce \textbf{TRAVL}, a fine-tuning recipe that combines a balanced training dataset with a trajectory-aware attention module to improve motion encoding and discrimination in VLMs. To evaluate physical reasoning more rigorously, we propose \textbf{ImplausiBench}, a benchmark of 300 videos (150 real, 150 generated) that removes linguistic biases and isolates visual-temporal understanding. Performance is reported both with gold-standard human judgments and stricter LLM-as-judge metrics. Together, TRAVL and ImplausiBench offer a unified framework for probing and improving physical plausibility in multimodal models, shedding light on a challenging and underexplored aspect of visual-temporal understanding.
\end{abstract}

\vspace{0.5em}
\end{@twocolumnfalse}
}]
\makeatother

\newpage

\section{Introduction}

Modern video generation models \cite{cogvideox2024, kong2024hunyuanvideo, runway2024, brooks2024worldsim} have achieved remarkable visual quality, yet they frequently produce sequences that violate intuitive physical laws—for example, objects may float, vanish, or morph in implausible ways. While such anomalies are easily detected by humans, quantitatively assessing physical realism in generated videos remains an open challenge~\cite{physics-iq, bansal2025videophy, meng2024towards, zhang2025morpheus}. Existing evaluation metrics like FVD~\cite{unterthiner2018accurate} and CLIPSIM~\cite{clip} prioritize perceptual similarity rather than physical plausibility. This raises a natural question: can video-language models (VLMs) be trained to serve as reliable judges of physical correctness in video? Motivated by the strong physics priors encoded in large language models~\cite{cherian2024llmphycomplexphysicalreasoning,shojaee2025llmsr}, we explore whether motion-aware visual grounding can enhance VLMs' ability to detect implausible dynamics.

Despite recent advances, VLMs still struggle to reason about physical plausibility and motion. Several studies highlight these limitations: MotionBench~\cite{hong2025motionbench} reports poor performance on fine-grained motion tasks involving multi-object interactions; Foresight-to-Forethought~\cite{wu2025foresight} shows that VLMs fail to predict outcomes in interactive physical scenarios; and Buschoff et al.~\cite{schulze2025testing} finds that fine-tuning on a narrow physics domain (e.g., falling blocks) fails to generalize to broader settings. Complementing these findings, recent benchmarks evaluating physical reasoning, such as PhysBench~\cite{physbench}, reveal that even the most capable models, including GPT-4o, perform well below human level~\cite{schulze2025testing, balazadeh2024synthetic}, particularly on tasks involving dynamic interactions. To compensate, hybrid systems like PhysAgent~\cite{physbench} inject symbolic or perceptual priors. Other efforts, such as Impossible Videos~\cite{bai2025impossible}, highlight the challenge of designing blind tests for implausibility detection, though structural and linguistic biases limit their use as a reliable evaluation set. In this work, we instead use such datasets as part of training material while shifting evaluation to more carefully constructed protocols.

Beyond benchmarks, architectural limitations also hinder physical reasoning. Current VLMs such as InternVideo~\cite{wang2022internvideo}, LLaVA-Video~\cite{llava-video}, Qwen2-VL~\cite{qwen-vl}, and Video-ChatGPT~\cite{video-chatgpt}—typically encode sparsely sampled frames independently via frozen image encoders like CLIP~\cite{clip} or SigLIP~\cite{siglip}. These representations are projected into the language model through simple adapters, discarding motion continuity and temporal context. As a result, these models often fail to recognize violations of physical laws, such as levitation, teleportation, or object morphing~\cite{bagad2023testoftime, bai2025impossible, physbench}. Addressing these shortcomings requires both better temporal grounding mechanisms and evaluation protocols that isolate genuine visual reasoning.

To address these challenges, we present both a fine-tuning recipe and an evaluation framework tailored to physical reasoning in video-language models. We introduce TRAVL (\textbf{TR}ajectory-\textbf{A}ware \textbf{V}ision-\textbf{L}anguage learning), a modular method that augments VLMs with motion-informed self-attention. TRAVL enhances visual encoding through two key mechanisms: (1) intra-frame spatial attention, which captures physically meaningful structure and relations within each frame—crucial for detecting anomalies like deformation, disappearance, or size inconsistencies; and (2) trajectory-aware temporal attention, which restricts inter-frame attention to follow sparse, object-level motion paths computed via CoTracker~\cite{karaev2024cotracker3}. This attention structure encourages the model to align visual tokens along both spatial structure and coherent motion, resulting in video representations that are more grounded in physical dynamics. TRAVL is lightweight and architecture-agnostic: it introduces no changes to the vision encoder or language model, and only fine-tunes a small number of attention and projection layers. Moreover, TRAVL is trained on a balanced dataset of plausible and implausible videos, ensuring robustness to distributional biases and improving generalization across diverse motion scenarios.

To rigorously evaluate physical reasoning capabilities, we introduce \textbf{ImplausiBench}, a benchmark explicitly designed to eliminate linguistic shortcuts and isolate visual-temporal understanding. ImplausiBench contains 300 videos (150 real, 150 generated), organized into paired plausible and implausible variants of the same scenario (sharing the same starting frame) and annotated with multiple-choice questions. Each question set was adversarially stress-tested in a \emph{blind evaluation protocol}, where off-the-shelf LLMs attempted to answer without viewing the video; whenever models exploited linguistic cues, we revised the multiple-choice answers until shortcut success was eliminated. In contrast, prior benchmarks such as Impossible Videos\cite{bai2025impossible} did not apply such blind testing, leaving them vulnerable to linguistic or positional biases. By construction, ImplausiBench ensures that progress reflects grounded video reasoning rather than surface-level patterns. Covering a broad spectrum of implausibility types including teleportation, levitation, deformation, duplication, and state changes, ImplausiBench serves as a high-fidelity diagnostic for evaluating whether VLMs truly understand physical plausibility in video.

\Needspace{5\baselineskip}
\noindent\textbf{Summary of Contributions.}
\begin{itemize}
\setlength\itemsep{0em}
\item We propose \textbf{TRAVL}, a modular fine-tuning recipe with trajectory-aware self-attention to enhance motion and physics understanding in VLMs.
\item We curate a balanced training dataset with plausible and implausible videos with a focus on physics reasoning. 
\item We propose \textbf{ImplausiBench}, a new benchmark of 300 videos that rigorously evaluates physical plausibility under both human and LLM-judge metrics.
\end{itemize}

\section{Related Work}
\begin{figure*}[t]
\centering
\includegraphics[width=0.9\linewidth]{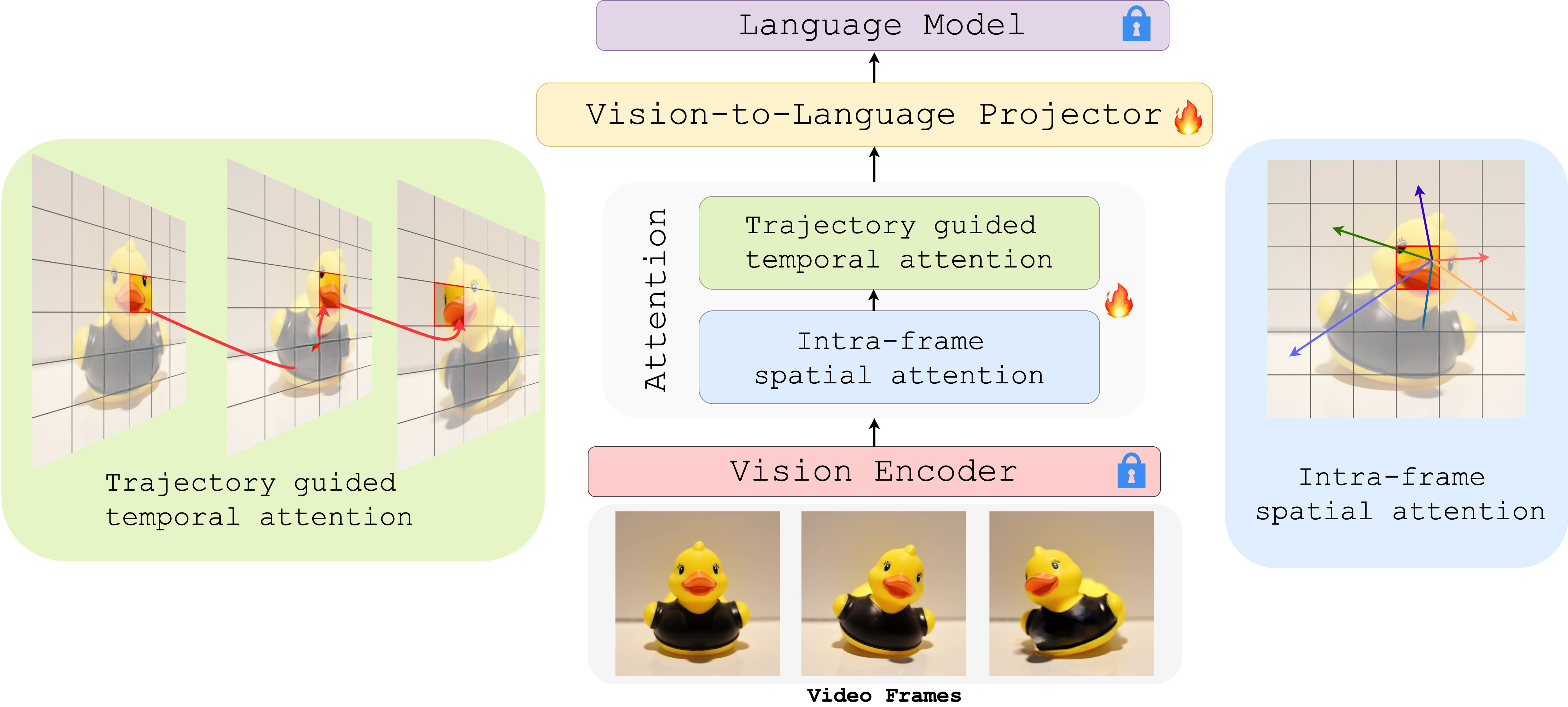}
\caption{\textbf{Overview of our proposed TRAVL framework.} Given input video frames, we apply a vision encoder followed by trajectory-aware masked self-attention, which integrates spatial and temporal context using patch trajectories tracked by CoTracker. The enriched features are projected into the language model's embedding space. Only the trajectory attention and vision-to-language projector are fine-tuned; the vision encoder and language model are kept frozen.}
\label{fig:method}
\end{figure*}

\subsection{Advancements and Limitations in Video-Language Models}

The development of Video-Language Models (VLMs) has been propelled by large-scale vision-language pretraining frameworks such as CLIP~\cite{clip}, ALIGN~\cite{jia2021scaling}, and SigLIP~\cite{siglip}.
These models form the backbone of more sophisticated video-capable architectures including InternVL~\cite{chen2024far}, Video-ChatGPT~\cite{video-chatgpt}, LLaVA-Video~\cite{llava-video}, and Qwen2.5VL~\cite{qwen2025qwen25technicalreport}, which perform well on standard video-language tasks such as captioning, retrieval, and Q/A. They have been evaluated on a variety of benchmarks including MMBench~\cite{liu2023mmbench}, MVBench~\cite{mvbench}, MTVQA~\cite{tang2024mtvqa}, MSRVTT-QA~\cite{xu2017videoqa}, MEGA-Bench~\cite{chen2024megabench}, VBench~\cite{huang2024vbench}, Video-Bench~\cite{ning2023videobench}, SEED-Bench~\cite{li2024seed}, and TempCompass~\cite{liu2024tempcompass}.
However, many of these benchmarks evaluate static understanding or treat frames independently, limiting insight into dynamic scene comprehension. Recent works~\cite{bagad2023testoftime, physbench} highlight that VLMs struggle with temporal coherence, motion continuity, and dynamic physical reasoning, motivating methods that inject stronger temporal grounding.

\subsection{Incorporating Trajectory-Based Temporal Modeling}

Trajectory-aware modeling has proven effective for capturing fine-grained motion in a variety of vision tasks. For instance, Motionformer~\cite{motionformer} uses trajectory attention to improve action recognition. FLATTEN~\cite{cong2023flatten}, pixel-aligned trajectory attention~\cite{xiao2024trajectory}, and VideoJAM~\cite{chefer2025videojam} enhance temporal consistency in video editing and generation, while OmnimatteZero~\cite{omnizero} improves training-free video inpainting using trajectory-aware attention. While these approaches showcase the benefits of motion-aware modeling, they are not designed to enhance physical plausibility reasoning in VLMs. Our work bridges this gap by integrating trajectory-guided attention into VLMs, enabling them to better track motion, detect temporal and spatial inconsistencies, and reason about physical implausibility.

\subsection{Evaluating Physical Reasoning in VLMs}

\paragraph{Benchmarks for Physical Reasoning.}
Several benchmarks have been developed to evaluate physical reasoning in both general AI systems and vision-language models.

InfLevel~\cite{weihs2022benchmarking} draws from infant cognition studies and uses a violation-of-expectation paradigm to evaluate whether models can detect core physical violations (e.g., continuity, solidity, gravity). It uses real-world and synthetic videos and is strictly diagnostic (no training is allowed). In contrast, \textbf{ImplausiBench} uses a multiple-choice Q/A format designed to probe causal and counterfactual reasoning, with adversarial distractors to prevent shortcut exploitation.

Melnik et al.~\cite{melnik2023benchmarks} provide a taxonomy of physical reasoning benchmarks based on reasoning type (descriptive, predictive, explanatory, counterfactual) and level of interaction. Key passive benchmarks include:
\begin{itemize}
    \item IntPhys~\cite{riochet2018intphys}: tests implausibility via frame prediction mismatch.
    \item CoPhy~\cite{baradel2019cophy}: evaluates prediction under modified initial conditions.
    \item CLEVRER~\cite{yi2019clevrer}: includes causal/counterfactual Q/A based on synthetic CLEVR videos.
    \item Physion~\cite{bear2021physion}: uses 3D simulations to test physical reasoning under gravity/collision.
\end{itemize}
While these benchmarks are valuable, they often rely on simplified synthetic data with minimal motion complexity. \textbf{ImplausiBench} extends this space by focusing on high-level physical plausibility in complex, real and generated videos with multiple objects and natural dynamics.

Physical Bongard Problems~\cite{weitnauer2012physical} test abstract physical concepts (e.g., stability, containment) through symbolic visual puzzles. Although the format differs from our visually grounded Q/A setting, the shared goal is interpretable physical understanding.

Virtual Tools~\cite{allen2020rapid} and PHYRE~\cite{bakhtin2019phyre} involve interactive tasks in 2D physics simulations. These are excellent for studying planning under physical constraints but are less applicable to VLMs, which operate in a passive video understanding setting without agent interaction.

\paragraph{Evaluating VLMs on Plausibility.}
PhysBench~\cite{physbench} introduces a comprehensive test suite for evaluating object dynamics and spatial interactions in real-world videos, while KiVA~\cite{yiu2024kiva} probes visual analogy-making in synthetic videos inspired by developmental psychology. However, neither benchmark explicitly targets implausible or counterfactual scenarios.

Impossible Videos~\cite{bai2025impossible} moves closer to our goal by evaluating whether models can detect physically, socially, or biologically implausible events via multiple-choice questions on generated videos. However, as we show in \cref{sec:implausibench}, their format is vulnerable to linguistic and positional biases that allow LLMs~\cite{openai2024gpt4o,yang2024qwen2} to succeed without robust visual grounding.
To address the limitations of prior benchmarks, we introduce \textbf{ImplausiBench}, a 300-video benchmark for evaluating physical plausibility in VLMs using paired plausible and implausible videos across diverse domains (e.g., cooking, sports, vehicles, shadows, reflections). Unlike earlier efforts, ImplausiBench:
\begin{itemize}
    \item targets both plausible and implausible temporal dynamics (e.g., levitation, teleportation, morphing, duplication) in real and generated videos,
    \item is rigorously designed to prevent shortcut exploitation via linguistic biases, and
    \item applies LLM-as-a-judge evaluation~\cite{gu2024survey,zheng2023judging,li2024generation} to normalize scoring across architectures, validated against full human evaluation.
\end{itemize}


\section{Method}

\begin{figure*}[t]
    \centering
    \includegraphics[width=\linewidth]{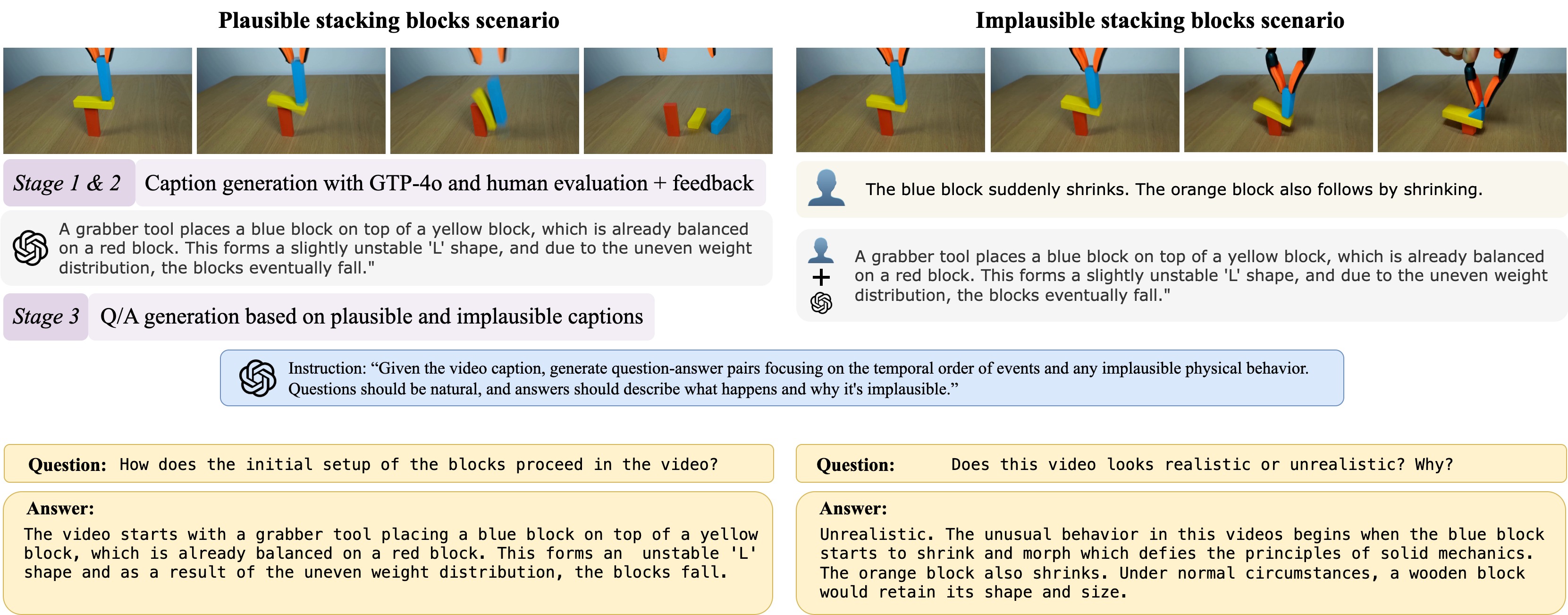}
    \caption{
    \textbf{Fine-tuning data pipeline.}
    Our dataset is built in three stages:
    \textbf{Stage 1 (Plausible Captioning):} GPT-4o generates initial captions for real (plausible) videos, verified by human reviewers.
    \textbf{Stage 2 (Feedback-Augmented Captioning):} Human annotators provide short temporal feedback for each implausible video, which is combined with the original real caption to create a complete description using GPT-4o.
    \textbf{Stage 3 (QA Generation):} Based on the final caption, GPT-4o produces temporally grounded question-answer pairs per video.
    This pipeline enables fine-grained supervision across a controlled set of plausible and implausible variants.
    }
    \label{fig:training-pipeline}
\end{figure*}

Understanding whether a video obeys the laws of physics often requires reasoning about both spatial configurations and object motion across time. For example, detecting implausibilities like objects hovering, teleporting, duplicating, or disappearing demands a joint understanding of structure and dynamics. To address this, we introduce \textbf{TRAVL}, a general-purpose fine-tuning recipe for pretrained video-language models. TRAVL incorporates trajectory-aware masked attention to enhance temporal and physical reasoning in VLMs. 

We first describe the attention mechanism itself (\cref{ssec:travl}), followed by its integration into existing VLM architectures (\cref{sec:model-integration}), and the fine-tuning dataset design that balances real and generated implausibilities (\cref{ssec:ftdata}).

\subsection{TRAVL}
\label{ssec:travl}

Modern VLMs typically begin with a vision encoder such as CLIP or SigLIP, which divides each frame into a grid of non-overlapping patches and maps each patch to a high-dimensional embedding. These visual embeddings are projected into the language model input space through lightweight adapters, enabling joint video-text reasoning. However, most VLMs encode each frame independently, discarding motion continuity and lacking mechanisms to capture spatial-temporal dynamics. As a result, they often fail to detect physically implausible motion patterns such as teleportation, deformation, or discontinuous trajectories.

\paragraph{Goal.}  
TRAVL introduces motion-aware attention into VLMs by combining intra-frame spatial attention with trajectory-guided temporal attention. Sparse patch trajectories, extracted using CoTracker~\cite{karaev2024cotracker3}, guide temporal connections, while spatial attention contextualizes patch structure within each frame. This dual attention design enables reasoning about both geometry (e.g., size, shape, occlusion) and continuity (e.g., persistence, gravity), without modifying the underlying vision or language backbones. 

We follow the patchification scheme of the vision encoder: e.g., $16\times16$ patches for CLIP (256 tokens per frame) or $27\times27$ for SigLIP (729 tokens per frame). Given $T$ frames, we extract patch embeddings $\mathbf{z}_{t,p}\in\mathbb{R}^d$, where $t=1..T$, $p=1..P$. To preserve layout and order, we add sine-cosine 2D spatial encodings and 1D temporal encodings prior to attention.

\paragraph{Intra-Frame Spatial Attention.}  
Self-attention across all patches $p=1..P$ within a frame $t$ models intra-frame structure:
\[
\mathbf{y}_{t,p} = \sum_{p'=1}^P \text{softmax}\!\left(\tfrac{\mathbf{q}_{t,p}^\top \mathbf{k}_{t,p'}}{\sqrt{d}}\right) \mathbf{v}_{t,p'}.
\]
The goal is to enhance detection of anomalies like duplication and deformation, aided by spatial positional encodings.

\paragraph{Patchwise Trajectory Masking.}  
To enforce temporal coherence, we track patch centers across frames and initialize new queries every $k$ frames for emerging objects. This produces a sparse binary mask $\mathbf{M}\in\{0,1\}^{TP\times TP}$ linking patches that share motion trajectories. The mask restricts temporal attention to physically plausible continuities (e.g., a rolling ball across time).

\paragraph{Trajectory-Guided Temporal Attention.}  
Temporal self-attention is restricted to valid links in $\mathbf{M}$:
\[
\mathbf{y}_i = \sum_{j:\mathbf{M}_{i,j}=1} \text{softmax}\!\left(\tfrac{\mathbf{q}_i^\top \mathbf{k}_j}{\sqrt{d}}\right)\mathbf{v}_j.
\]
This enforces object persistence, enabling detection of implausibilities like teleportation or sudden morphing. Following both spatial and temporal attention, enriched patch embeddings are projected to the language space through a learnable adapter. The vision encoder and language model remain frozen; only TRAVL’s attention and projection modules are trained. Figure \ref{fig:method} shows an overview of TRAVL's main components.

\subsection{Model Integration: TRAVL Across Architectures}
\label{sec:model-integration}

We validate TRAVL on two representative VLMs, demonstrating its modular integration in both pooled and dense token settings.

\paragraph{Video-ChatGPT.}  
Video-ChatGPT pools $256$ CLIP patch tokens from $100$ frames into temporal and spatial summaries before projection. With TRAVL, we replace pooling with intra-frame spatial attention and trajectory-guided temporal attention over sparse CoTracker masks. The resulting enriched tokens are passed through a lightweight projection. Only these new modules are trained; CLIP and the LLM stay frozen.

\paragraph{LLaVA-NeXT.}  
LLaVA-NeXT encodes $64$ frames via SigLIP into $729$ patch tokens per frame. The original spatial pooling is replaced with TRAVL’s spatial and chunked temporal attention (e.g., 4–16 frame windows), guided by sparse trajectories. The attended features are fused, pooled, and projected. TRAVL thus preserves input-output format while injecting motion-awareness. Our ablations confirmed that both spatial-only and temporal-only modules improved plausibility detection, but the full TRAVL design yielded the best results.

\subsection{Fine-tuning Dataset}
\label{ssec:ftdata}

To train TRAVL-equipped VLMs, we curate a dataset that balances plausible and implausible videos while retaining broad video-language coverage. Our design emphasizes natural failure cases from generative models and balanced question types to ensure physically grounded learning. Figure~\ref{fig:training-pipeline} shows our training data generation pipeline.

\begin{figure*}[t]
    \centering
    \includegraphics[width=\linewidth]{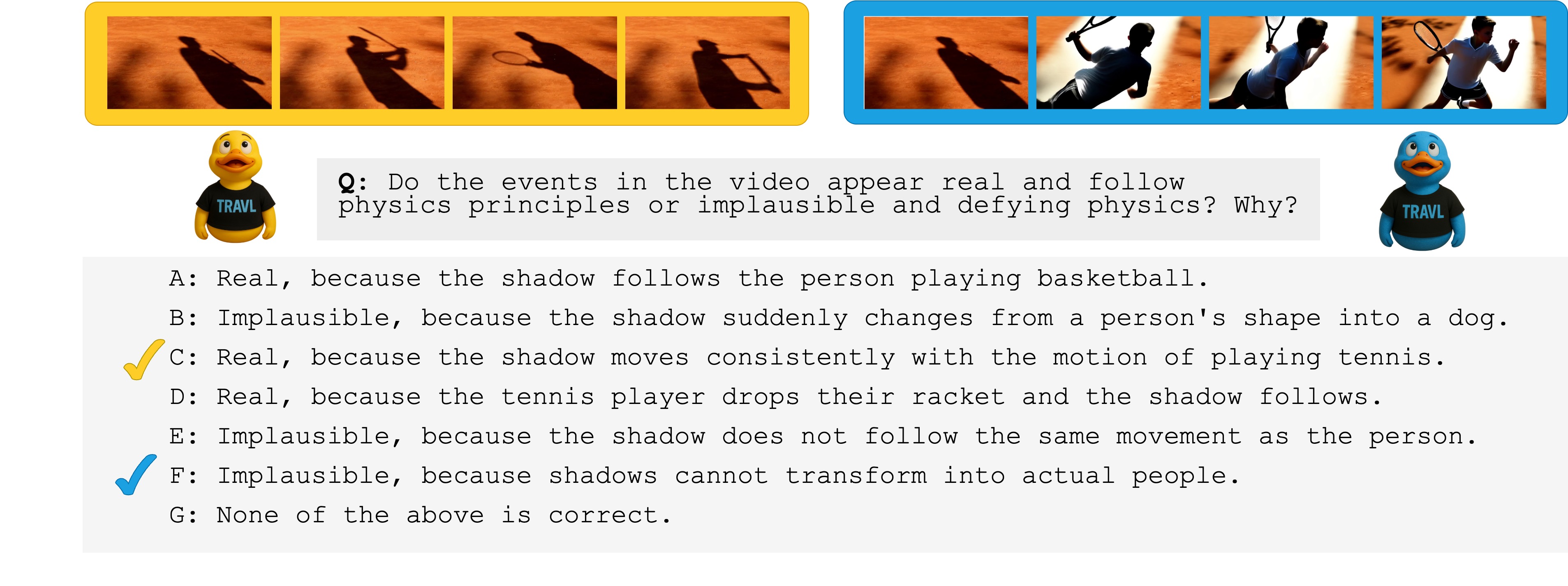}
    \caption{\textbf{Example from ImplausiBench.} For each scenario, we include both a real (plausible) and a generated (implausible) video that share the same initial scene and visual style. Each pair is annotated with a shared multiple-choice question containing three plausible, three implausible, and one ``None of the above'' option. The correct answer depends on which version of the video is shown—ensuring that models must ground their predictions in visual-temporal evidence rather than language alone.}
    \label{fig:impbench-sample}
\end{figure*}

\paragraph{Physics-IQ Scenarios with Synthetic Violations.}
We begin with 66 base scenarios from Physics-IQ~\cite{physics-iq}, each captured from three views (198 plausible videos). Using Runway, Pika, Sora, Kling, and Luma, we generate 894 variants conditioned on captions and first frames. 
Unlike prior works that induce violations through prompt engineering, we do not modify prompts to force implausibility. Instead, we capture natural failure cases of diffusion models. Human annotators review every generation, discarding approximately 70 cases where outputs were either plausible or static. 

For each retained implausible clip, annotators provide structured temporal descriptions of the violations as they occur (e.g., ``duck disappears midair, then reappears in a different location''). These fine-grained annotations are used to guide GPT-4o in producing detailed captions and generating balanced Q/A pairs. This annotation pipeline ensures that implausible events are faithfully represented.

\paragraph{Impossible Videos.}
Since the multiple-choice format of Impossible Videos~\cite{bai2025impossible} has been shown to admit language-only shortcuts (see Table~\ref{tab:blindtest}), we do not use it as an evaluation benchmark. Instead, we repurpose 535 clips from its ``Physics'' category as training material. To broaden the coverage of implausibility types, we also generate 92 additional clips with Pika, applying the same human verification and temporally grounded annotation pipeline described above. This ensures consistency in how implausible events are identified, described, and paired with balanced Q/A prompts.

\paragraph{Real-World QA from Video-ChatGPT.}
To maintain generalization beyond synthetic distortions, we include 1,763 diverse clips from the Video-ChatGPT training set, paired with their original QA annotations. We filter out long clips ($>$800 frames) to ensure patch trajectories remain temporally meaningful.

\paragraph{Dataset Statistics.}
The final dataset contains \textbf{3,482 videos} and \textbf{19,708 QA pairs}.  
A key design principle is balancing the \emph{types of questions} across both plausible and implausible videos to avoid \textit{dataset skew}. In particular, plausibility-style questions (e.g., ``Does the video look real or implausible?'') are deliberately posed not only for implausible clips but also for real ones. By ensuring that every QA type is mirrored across both categories, we prevent models from exploiting correlations between question form and video class. This balanced QA distribution requires models to ground their answers in visual evidence rather than annotation patterns. Additional details and examples are provided in Supplementary section ~\ref{supp:ftdataset}.

\section{ImplausiBench}
\label{sec:implausibench}

In this section, we present \textbf{ImplausiBench}, a diagnostic benchmark designed to test whether video-language models can detect physically implausible events using visual-temporal cues alone. It consists of paired plausible and implausible videos constructed to minimize language-only shortcuts and isolate grounded physical reasoning.

\paragraph{Benchmark Construction.}
ImplausiBench comprises 150 real-world videos depicting physically plausible scenes. For each, we synthesize an implausible counterpart using state-of-the-art diffusion-based video models (e.g., Pika~\cite{pika2024}, Runway~\cite{runway2024}, Kling~\cite{kling2024}, CogVideo~\cite{cogvideox2024}, LTX~\cite{LTX}, Pyramid-Flow~\cite{pyramid}), conditioned on a GPT-4o-generated caption and the first frame of the real video. If the generated result remains plausible after manual inspection, we regenerate until a clear physical violation is introduced.

The resulting videos capture a broad spectrum of implausibility types, loosely grouped into six categories: \textit{motion anomalies} (e.g., levitation, reversal), \textit{object continuity violations} (e.g., teleportation, disappearance), \textit{structural transformations} (e.g., deformation, splitting), \textit{unnatural interactions} (e.g., passing through solids), \textit{appearance shifts} (e.g., sudden color or size changes), and \textit{implausible state changes} (e.g., self-filling, melting). These failure modes align with key principles in intuitive physics and reflect typical breakdowns in generative video models.

\paragraph{Multiple Choice Format.}
Each plausible–implausible video pair is annotated with a single shared multiple-choice question (MCQ) containing \textbf{seven answer options}: three describing plausible outcomes, three describing implausible ones, and one \emph{``None of the above''} option. 
Unlike prior benchmarks that rely heavily on automated generation, all answer options in ImplausiBench are manually curated by annotators to ensure clarity, precision, and grounding in the visual content. 
To guard against shortcut exploitation, we perform a \emph{blind test validation}: off-the-shelf LLMs are asked to answer the MCQs without access to the video. Whenever models succeed above chance by exploiting linguistic or positional patterns, we revise the answer set until such shortcuts are eliminated. 
This rigorous process makes ImplausiBench resistant to language-only biases—a key limitation of datasets like Impossible Videos—and ensures that correct answers depend on visually grounded reasoning. 
Due to the intensive manual effort required to design, review, and validate each question, the benchmark is intentionally limited to 300 videos, prioritizing annotation quality over scale.

\newcolumntype{L}[1]{>{\raggedright\arraybackslash}p{#1}}

\begin{table}[t]
\centering
\caption{Blind test multiple-choice accuracy (no video shown). Random chance is 14.3\% for ImplausiBench (7 options) and 20\% for Impossible Videos (5 options).}
\label{tab:blindtest}
\scriptsize   
\setlength{\tabcolsep}{3pt} 
\renewcommand{\arraystretch}{1.1} 
\begin{tabular}{lL{2.0cm}L{2.0cm}L{2.0cm}}
\toprule
\textbf{Model} & \textbf{Impossible Videos} & \textbf{ImplausiBench (implausible)} & \textbf{ImplausiBench (plausible)} \\
\midrule
GPT-4o & 51.2\% & 22\% & 21.3\% \\
Qwen2.5-7B & 46\% & 20\% & 18.6\% \\
Random & 20\% & 14.3\% & 14.3\% \\
\bottomrule
\end{tabular}
\end{table}

\paragraph{Comparison to Existing Benchmarks.}
We apply the blind test protocol to both ImplausiBench and the Physics category of the Impossible Videos benchmark~\cite{bai2025impossible}. In Impossible Videos, each MCQ presents one plausible and four implausible answers, with prompts that implicitly bias models toward selecting an implausible option, even when no video is shown. This evaluation format permits shortcut exploitation: as \Cref{tab:blindtest} shows, GPT-4o and Qwen2.5 achieve well above chance-level accuracy without visual input. ImplausiBench avoids this pitfall by balancing plausible and implausible choices, including a ``None of the above'' option, and filtering out easy distractors. As a result, blind-test accuracy drops closer to chance. 

\paragraph{Why We Do Not Use Impossible Videos for Evaluation.}
Since Impossible Videos allows models to exploit linguistic and positional biases without grounding in visual evidence, we do not rely on it for evaluation. Instead, we repurpose its videos for training, where implausible content remains valuable for supervision. ImplausiBench, by contrast, enforces stricter correctness (models must succeed on both plausible and implausible versions of each scenario) and is adversarially constructed to resist such shortcuts. This makes it a more rigorous benchmark for assessing physical reasoning and visual grounding in VLMs.

\section{Results}

\begin{table}[t]
\centering
\caption{Evaluation on \textbf{ImplausiBench}, split into \textbf{Implausible} and \textbf{Real} subsets (150 videos each). 
Numbers are accuracies in \%. As the gold standard, we report the \emph{Human} evaluation metric, based on user judgments of correctness for each VLM output. 
For comparison, we also report the \emph{LLM-judge} evaluation, which provides a stricter automatic assessment.}

\scriptsize
\begin{tabular}{lcccc}
\toprule
\textbf{Model} & \multicolumn{2}{c}{\textbf{Implausible (150)}} & \multicolumn{2}{c}{\textbf{Real (150)}} \\
& \textbf{Human} & \textbf{LLM} & \textbf{Human} & \textbf{LLM} \\
\midrule
\multicolumn{5}{c}{\textit{Proprietary}} \\
\midrule
GPT-4o              & 32.7 & 21.3 & 84.7 & 64.0 \\
Gemini 2.5 Pro      & 41.3 & 29.3 & 100.0 & 78.0 \\
\midrule
\multicolumn{5}{c}{\textit{Open-Source}} \\
\midrule
Qwen2.5VL            & 18.7 & 12.0 & 96.7 & 74.7 \\
InternVideo2.5         & 12.7 & 4.7 & 92.7 & 76.0 \\
Video-ChatGPT Pre-trained   & 0.0 & 0.0 & 72.0 & 55.3 \\
Video-ChatGPT SFT   & 6.0 & 2.7 & 39.3 & 26.0 \\
\textbf{Video-ChatGPT TRAVL} & \textbf{12.0} & \textbf{7.3} & \textbf{42.7} & \textbf{31.3} \\
LLaVA-NeXT Pre-trained        & 3.3 & 2.7 & 98.7 & 62.7 \\
LLaVA-NeXT SFT           & 34.0 & 22.0 & 45.3 & 23.3 \\
\textbf{LLaVA-NeXT TRAVL}         & \textbf{52.7} & \textbf{28.7} & \textbf{47.3} & \textbf{31.3} \\

\bottomrule
\end{tabular}
\label{tab:implausibench}
\end{table}

\paragraph{Evaluation Protocol.}
We evaluate models on \textbf{ImplausiBench}, split into \textit{Generated} and \textit{Real} subsets (150 videos each). For each subset we report accuracy in \% under two metrics: (i) a \textbf{Human} metric, where annotators watched each video and judged whether the model’s caption correctly described it, and (ii) an \textbf{LLM-judge} metric. To ensure comparability across models with open-ended vs.\ multiple-choice formats, we adopt an LLM-as-a-judge protocol~\cite{gu2024survey,zheng2023judging,li2024generation}. Each model answers the same open-ended prompt (\textit{``Do the events in the video appear to be real, following physics principles, or are they implausible? Why?''}); GPT-4o then maps the response to the benchmark’s multiple-choice options. The LLM-judge is explicitly instructed to be strict: if an answer is partially correct or omits critical details about the violation, it often defaults to the \emph{``None of the above''} option rather than granting partial credit by picking the closest answer to the caption. This conservative scoring reduces the risk of inflating model performance but also leads to lower absolute scores compared to human judgment. To anchor results, human annotators reviewed every model output, providing the \textbf{gold standard} Human metric. Importantly, while the LLM-judge yields stricter scores, the relative trends between models are preserved, making it a reliable and cautious proxy for large-scale evaluation.

\paragraph{Scoring on ImplausiBench.}
We award credit separately on the \textit{Generated} and \textit{Real} subsets (\Cref{tab:implausibench}). This design makes performance on synthetic violations (Generated) and naturally plausible videos directly comparable, while separating human-verified correctness from automated judging. 
dcjbfjuhcbguncgnjrihgcckcigjfvtv

\paragraph{TRAVL Improves Implausibility Detection.}
Across both backbones, adding \textbf{TRAVL} yields consistent gains on the \textit{Generated} subset. On the \textit{Real} subset, pretrained models can appear stronger, but this is misleading: they achieve high scores by defaulting to “plausible” predictions while failing almost entirely on implausible cases. A fairer comparison is against the \textbf{SFT} baseline, which is trained on the same data distribution but without TRAVL. Relative to SFT, TRAVL improves performance on both subsets under both Human and LLM-judge metrics. For instance, LLaVA-NeXT with TRAVL outperforms SFT by $18.7\%$ on implausible videos and $2.0\%$ points on real ones (Human metric). Similar improvements hold for Video-ChatGPT. These results confirm that spatial and trajectory-guided temporal attention modules strengthens motion grounding and detection of physical violations, while preserving general plausibility understanding.

\subsection{Ablation Studies}
\label{sec:ablation}

To better understand the contributions of TRAVL’s components, we ablate its two attention modules: spatial self-attention and trajectory-guided temporal attention. 
Both variants are trained with the same settings as TRAVL but with only one component active at a time. 
This reveals whether improvements in implausibility detection arise primarily from intra-frame spatial grounding or trajectory-guided temporal attention. The results are shown in Table ~\Cref{tab:ablation}.

\begin{table}[t]
\centering
\caption{\textbf{Ablation on LLaVA-NeXT.} 
Evaluation on ImplausiBench. 
Numbers are accuracies in \%.}
\scriptsize
\begin{tabular}{lcccc}
\toprule
\textbf{Model} & \multicolumn{2}{c}{\textbf{Implausible (150)}} & \multicolumn{2}{c}{\textbf{Real (150)}} \\
& Human & LLM & Human & LLM \\
\midrule
Pretrained LLaVA-NeXT & 3.3 & 2.7 & 98.7 & 62.7 \\
LLaVA-NeXT SFT & 34.0 & 22.0 & 45.3 & 23.3 \\
Temporal-only Attention & 46.0 & 24.0 & 41.3 & 22.0 \\
Spatial-only Attention & 42.7 & 26.7 & 48.7 & 30.7 \\
\textbf{TRAVL (Ours)} & \textbf{52.7} & \textbf{28.7} & \textbf{47.3} & \textbf{31.3} \\
\bottomrule
\end{tabular}
\label{tab:ablation}
\end{table}

\paragraph{Findings.}
Both spatial-only and temporal-only variants improve over supervised fine-tuning, but neither matches the full TRAVL model. 
This indicates that spatial and temporal attention provide complementary benefits: spatial attention enhances detection of implausible structures (e.g., overlaps, deformations), while temporal attention improves motion continuity tracking. 
Together, they yield the strongest overall gains in plausibility reasoning.

\paragraph{Binary Classification Results.}
We also evaluate models in a binary plausibility classification setup, where the task is to label each video as plausible or implausible. This metric does not probe reasoning quality, but provides a complementary view of discrimination ability. As shown in ~\Cref{tab:binary}, TRAVL improves implausibility detection while maintaining plausible video accuracy, with ablated variants again performing between SFT and full TRAVL.

\begin{table}[h]
\centering
\caption{Binary classification accuracy (\%) of LLaVA-NeXT models on \textbf{ImplausiBench}.}
\label{tab:binary}
\scriptsize
\begin{tabular}{lcc}
\toprule
\textbf{Model} & \textbf{Real (Plausible)} & \textbf{Implausible} \\
\midrule
LLaVA-NeXT Pre-trained & 98.7 & 10.0 \\
LLaVA-NeXT SFT & 45.3 & 83.3 \\
LLaVA-NeXT Temporal-only Attention & 52.0 & 82.7 \\
LLaVA-NeXT Spatial-only Attention & 53.3 & 84.7 \\
\textbf{LLaVA-NeXT TRAVL (Ours)} & \textbf{57.3} & \textbf{84.0} \\
\bottomrule
\end{tabular}
\end{table}

\section{Limitations and Future Work}
\label{sec:limitations}

While TRAVL advances temporal modeling and physical plausibility detection in VLMs, some limitations remain. Our fine-tuning dataset is modest in size and limited in diversity relative to real-world video content; expanding to broader categories of physical implausibility and scenarios, potentially via automated generation pipelines or simulation environments, could improve generalization. TRAVL also depends on externally generated patch trajectories, introducing computational overhead and sensitivity to visual artifacts such as occlusion or blur, and integrating learned or differentiable tracking directly into the model may improve robustness. In dense-input settings (e.g., LLaVA-NeXT), temporal attention is applied over short video chunks (4–16 frames) to maintain tractability, which limits long-range reasoning; future work could explore memory-efficient attention to enable full-sequence modeling. Finally, our mediated evaluation relies on GPT-4o to judge model outputs, introducing a dependency on another language model’s interpretation. Despite these limitations, TRAVL provides a lightweight and extensible strategy for integrating temporal structure into VLMs, and ImplausiBench offers a high-fidelity benchmark for assessing visual-temporal physical understanding.

\section{Conclusion}

We introduced \textbf{TRAVL}, a trajectory-aware fine-tuning framework that improves physical reasoning in VLMs by integrating spatial and trajectory-aware temporal attention and plausibility supervision. TRAVL enables pretrained VLMs to better detect implausible motion patterns with minimal modifications to their vision or language backbones. We demonstrated its effectiveness on both Video-ChatGPT and LLaVA-NeXT, showing consistent gains in physical plausibility judgment.

To enable more rigorous evaluation, we proposed \textbf{ImplausiBench}, a benchmark designed to eliminate linguistic shortcuts and isolate visual-temporal understanding. Our blind test protocol confirms that ImplausiBench is significantly more robust to shortcut exploitation than existing benchmarks such as Impossible Videos, offering a clearer signal of grounded physical reasoning.

\section{Acknowledgment}
This research was partially funded by the Ministry of Education and Science of Bulgaria (support for INSAIT, part of the Bulgarian National Roadmap for Research Infrastructure). The authors would like to thank Raha Ahmadi for supporting this project by helping with ImplausiBench dataset.

\newpage
{
    \small
\bibliographystyle{plain}
    \bibliography{neurips_2025}
}

\newpage
\appendix

\clearpage

\begin{center}
\LARGE \textbf{TRAVL: Supplementary Material}
\end{center}

This supplementary material provides expanded details supporting our main paper. We begin by describing the GPT-4o-based evaluation protocol we use to score open-ended VLM responses against multiple-choice ground truth. We then present qualitative visualizations from both the Impossible Videos and our newly proposed ImplausiBench datasets, highlighting model successes and failures across different training stages. Quantitative results are further broken down to analyze tradeoffs between plausibility sensitivity and implausibility detection. We outline the structure of our fine-tuning dataset and provide the prompt design used for generating temporal and physics-based QA pairs in Pseudocode \ref{lst:qagen}. In Section~\ref{supp:implausibench}, we detail the construction of ImplausiBench. Finally, we document TRAVL’s model architecture, training setup, practical observations, and how it compares with prior physical reasoning benchmarks. To view the example videos referenced throughout, please open \texttt{result\_viewer.html} in the supplementary folder.

\section{Results in Detail}
\label{supp:results}

\paragraph{LLM-as-a-Judge Evaluation.}
To evaluate whether a vision-language model’s (VLM’s) open-ended response corresponds to the correct multiple-choice answer, we adopt an LLM-as-a-judge protocol with GPT-4o. Each VLM is first prompted to provide an open-ended explanation of physical plausibility. GPT-4o then receives this explanation together with the corresponding multiple-choice question and candidate options, and is instructed to select the option that best matches the VLM’s reasoning. Importantly, GPT-4o is not told the ground truth during evaluation; its role is to strictly map the VLM’s free-form output to one of the benchmark’s predefined answers. 

We validated this judge protocol using blind probes (no video input) to ensure it does not exploit language bias. To prevent partial credit, we also include a ``None of the above'' fallback option in every question. In Pseudocode \ref{lst:judge} we show the exact function used to construct judge prompts.

\begin{figure*}[t]
\begin{lstlisting}[language=Python, caption={LLM-as-judge prompt},label={lst:judge}]
def ask_gpt(client, model, question, options, caption, video_name, attempt_limit=3):
    options_text = "\n".join([f"{k}. {v}" for k, v in options.items()])
    prompt_messages = [
        {
            "role": "user",
            "content": (
                "You are a reasoning assistant evaluating the output of a "
                "video-language model.\n\n"
                "The VLM model has watched a video and described the video as:\n"
                f"{caption}\n\n"
                f"Based on the above answer and analyzing its reasoning to the "
                f"question of: {question}, select which of the following "
                "multiple-choice options best matches the model's reasoning.\n"
                "Your judgment should be based only on the VLM's output.\n"
                "Respond with the letter of the best matching option.\n\n"
                f"Options:\n{options_text}"
            )
        }
    ]
\end{lstlisting}
\end{figure*}

\paragraph{Visualizations of VLM Outputs.} 
Figure~\ref{fig:implausibench-samples} showcases qualitative examples from both Video-ChatGPT and LLaVA-NeXT backbones, along with their supervised and TRAVL fine-tuned variants. The examples span four implausible videos from the Impossible Videos benchmark and four plausible–implausible video pairs from our ImplausiBench benchmark. We highlight diverse model behaviors, including cases where different versions succeed (check mark) or fail (cross) in answering the open-ended prompt: \textit{“Do the events in the video appear to follow physics principles or not? Why?”} For full visualizations and model outputs, please view \texttt{result\_viewer.html} in the supplementary folder.

\begin{figure*}[t]
\centering
\includegraphics[width=0.9\textwidth]{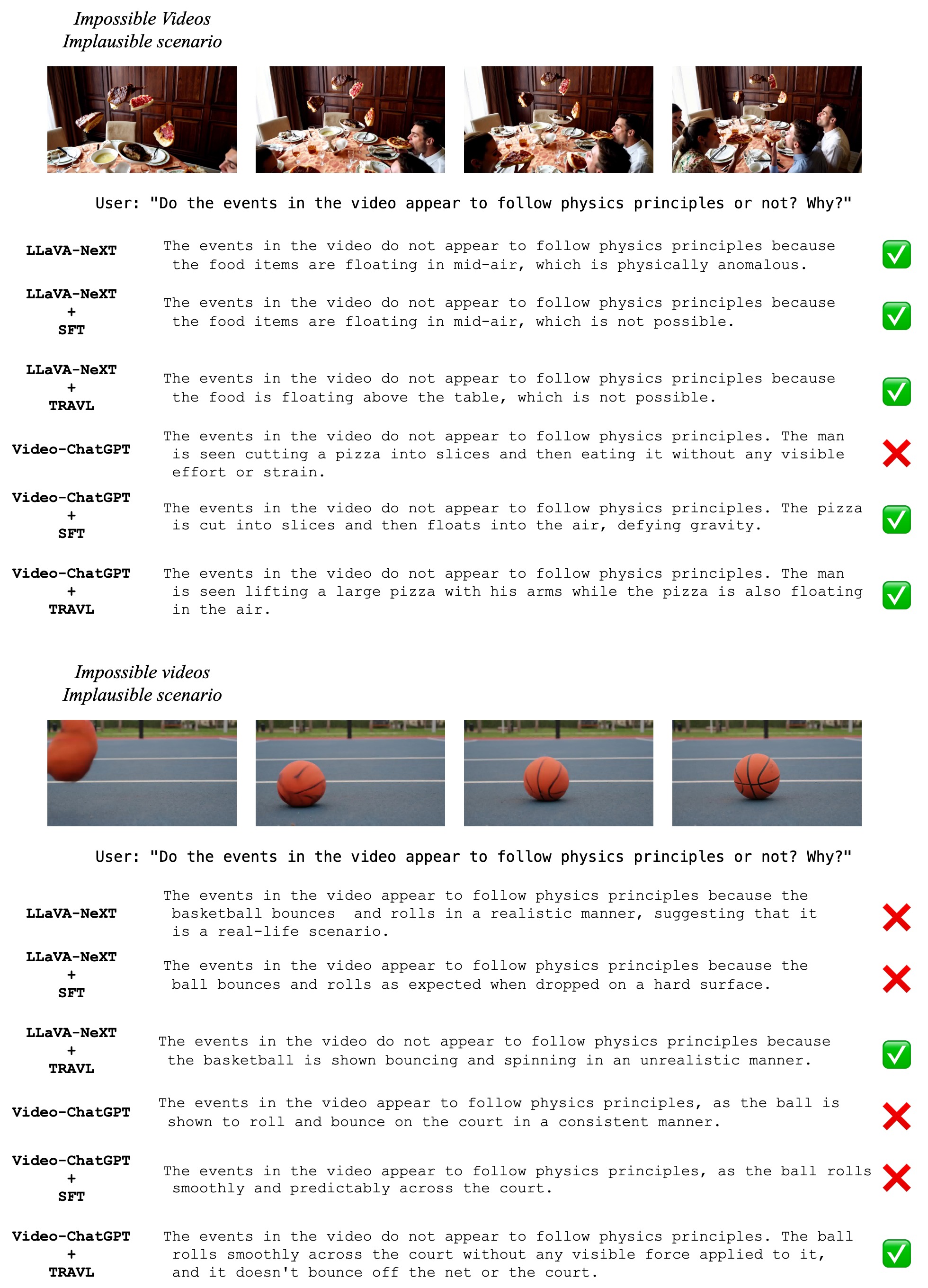}
\end{figure*}

\begin{figure*}[t]
\centering
\includegraphics[width=0.9\textwidth]{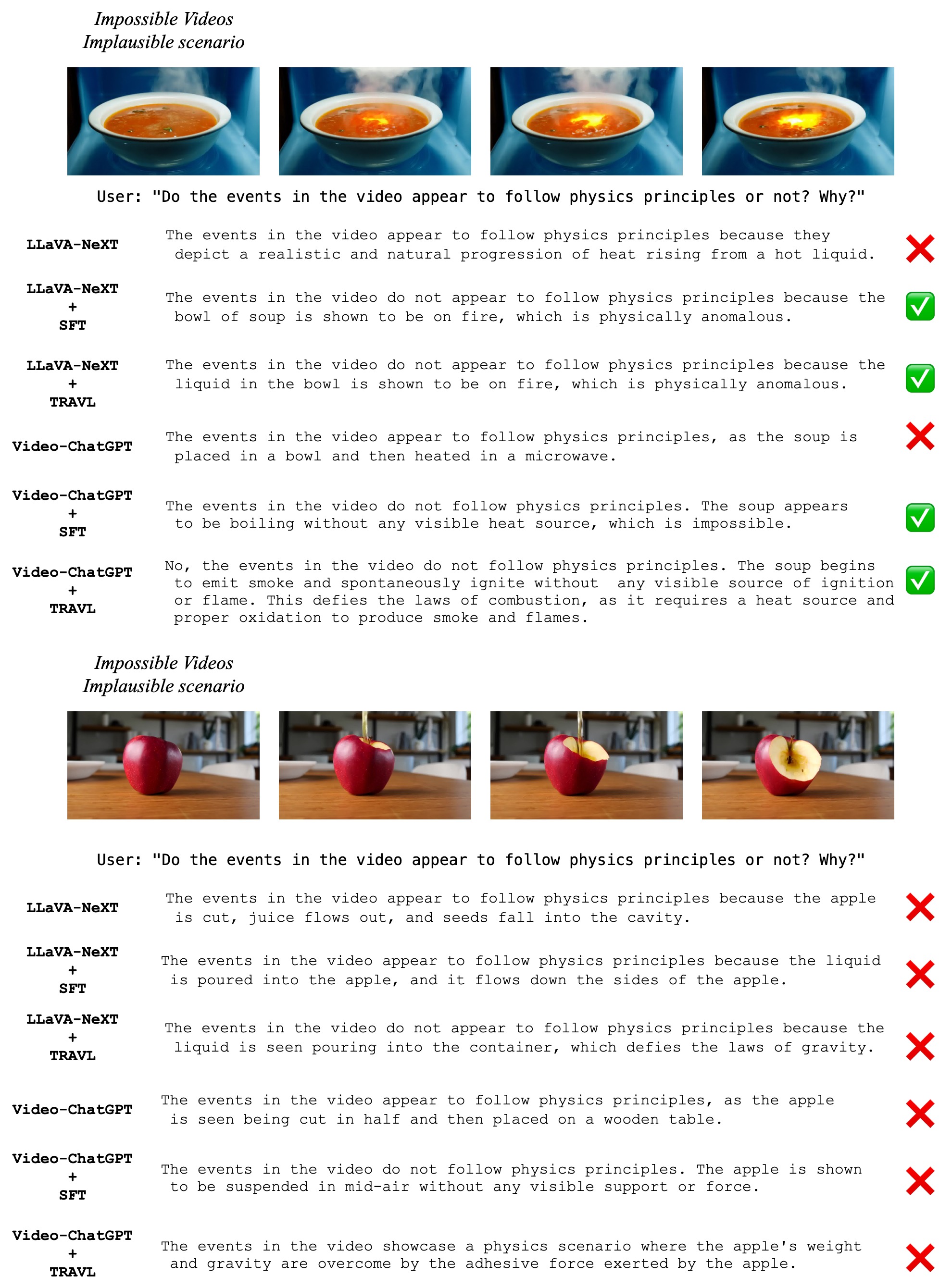}
\end{figure*}

\begin{figure*}[t]
\centering
\includegraphics[width=0.9\textwidth]{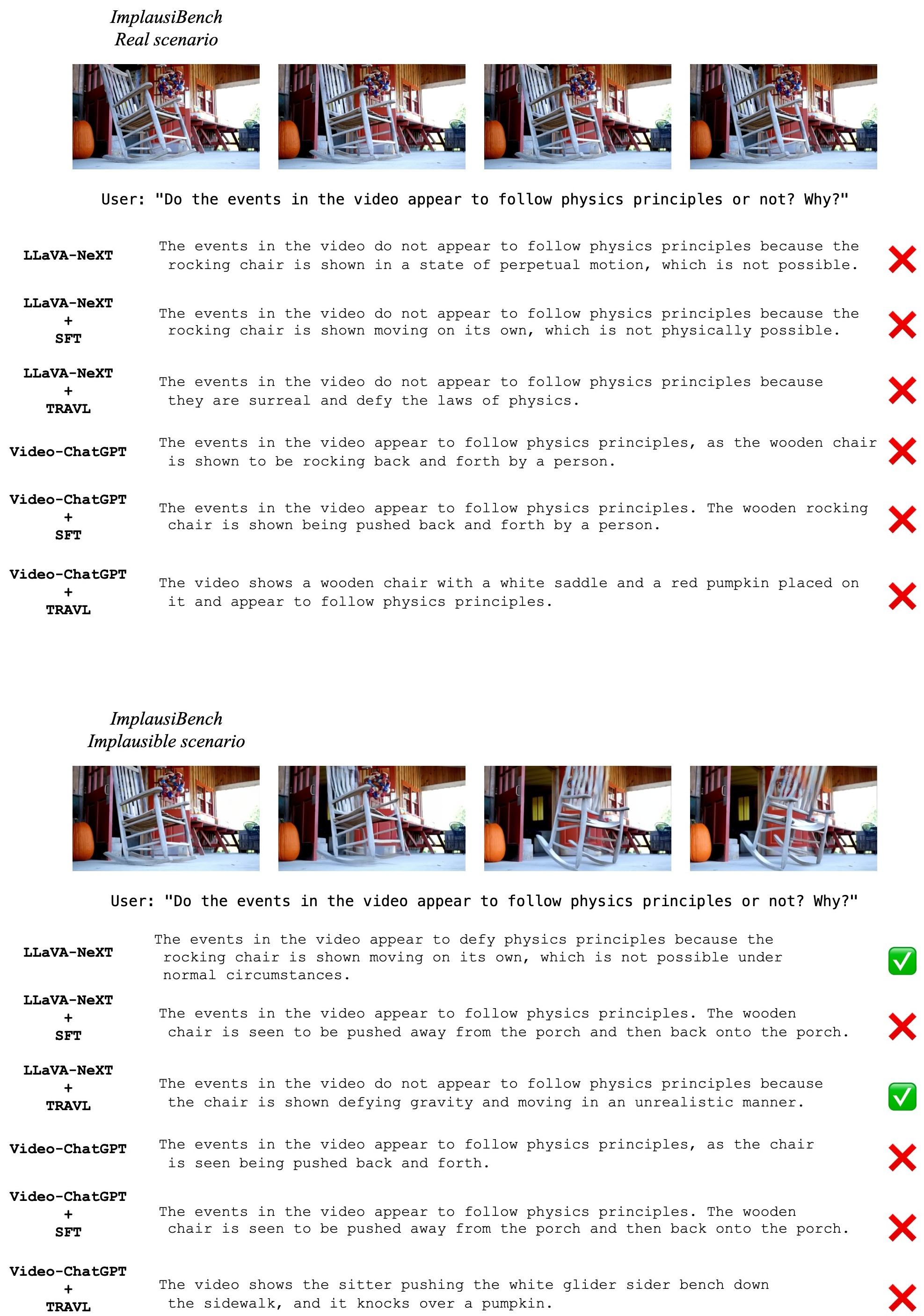}
\end{figure*}

\begin{figure*}[t]
\centering
\includegraphics[width=0.9\textwidth]{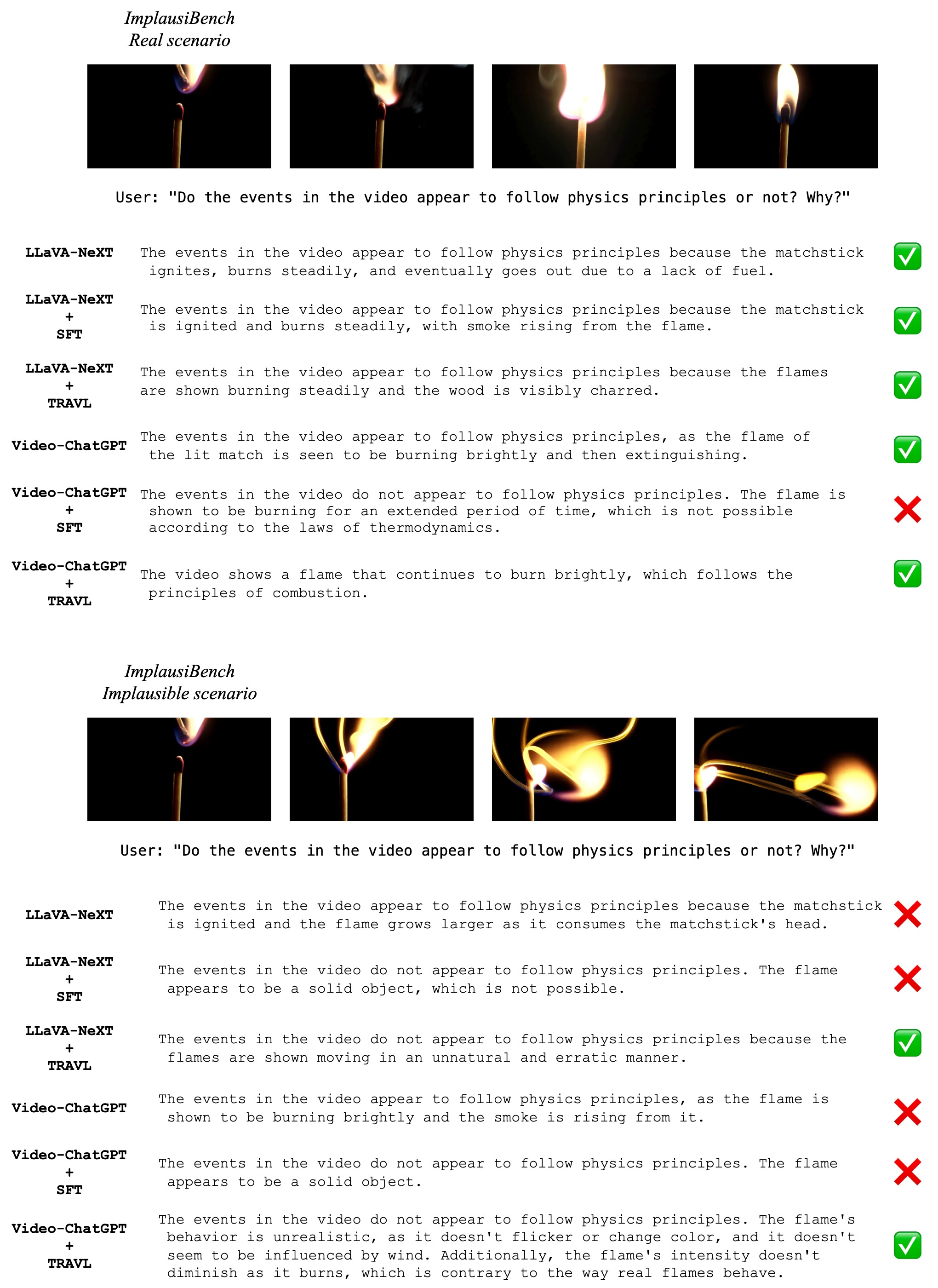}
\end{figure*}

\begin{figure*}[t]
\centering
\includegraphics[width=0.9\textwidth]{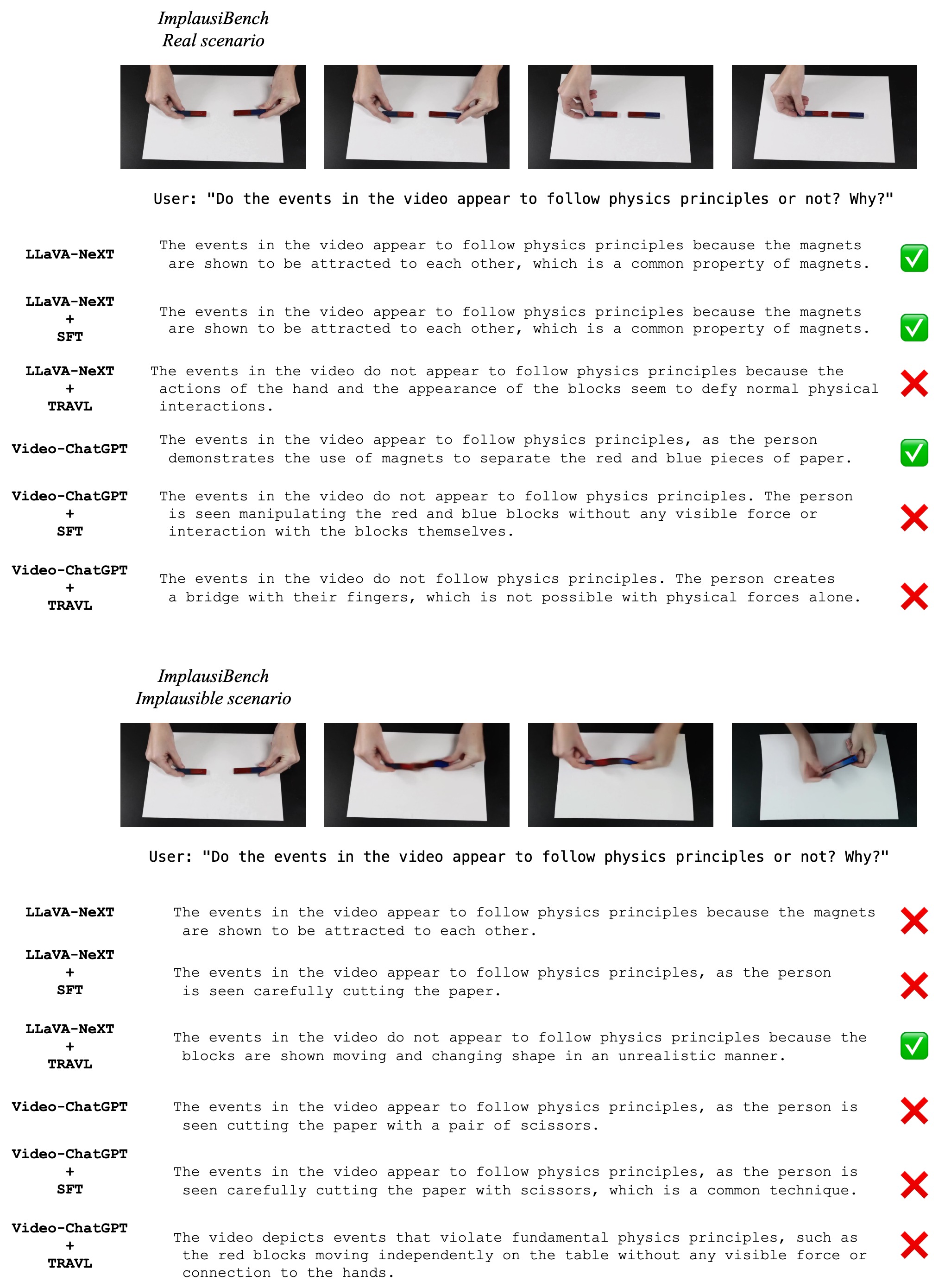}
\end{figure*}

\begin{figure*}[t]
\centering
\includegraphics[width=0.9\textwidth]{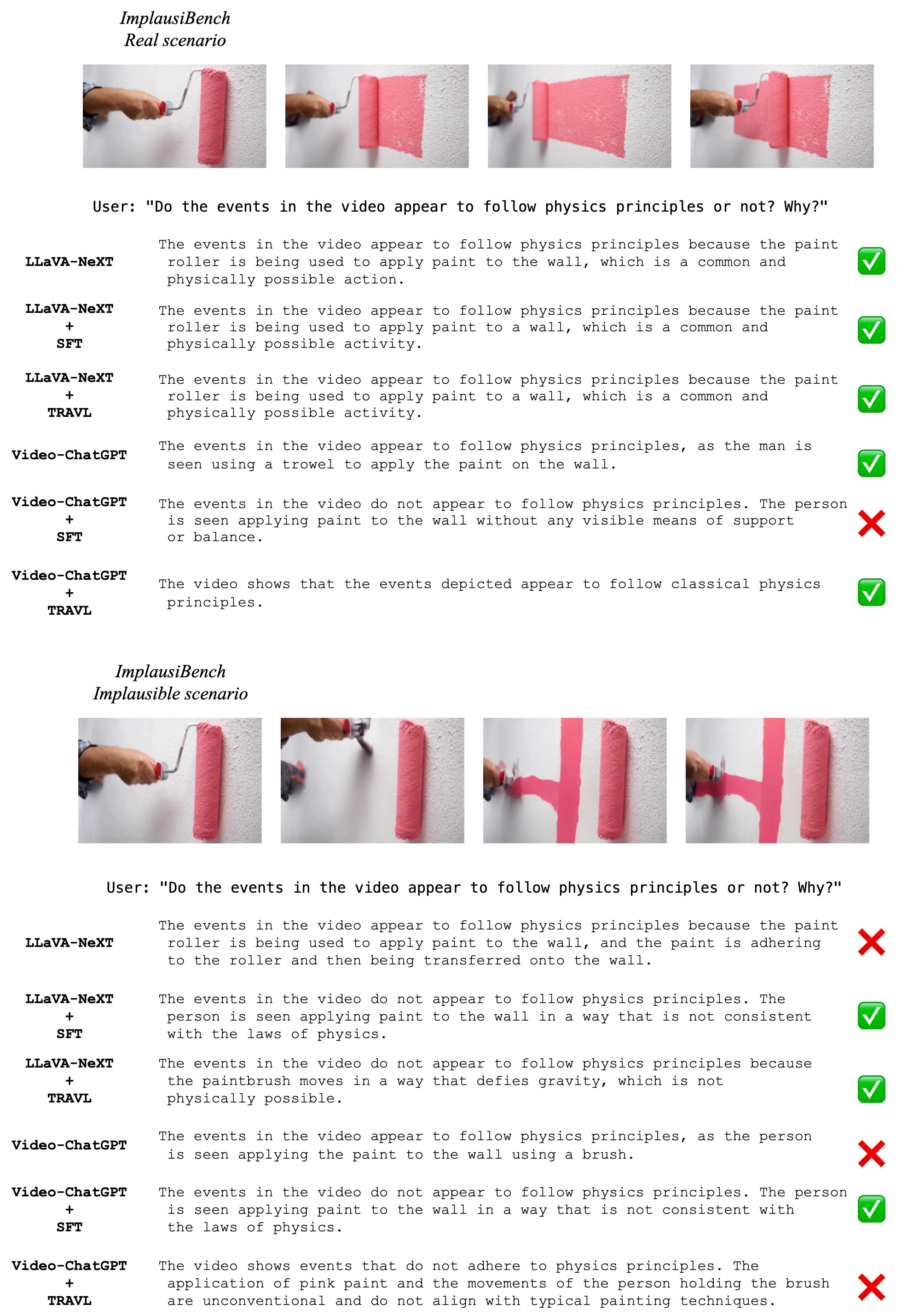}
\vspace{1em}
\captionof{figure}{\textbf{Qualitative examples from TRAVL.} 
The first two pages show frames from Impossible Videos, while the remaining illustrate plausible and implausible variants from ImplausiBench. These examples were selected to showcase representative successes (check mark) and failures (cross) across different models, as identified through manual inspection.}

\label{fig:qual-examples}
\end{figure*}

\paragraph{Understanding the Plausibility–Implausibility Tradeoff.}
Table~\ref{tab:implausibench} reports accuracy as a percentage of correct predictions out of 150 videos for both plausible and implausible variants in ImplausiBench. Untuned models such as Video-ChatGPT and LLaVA-NeXT show high accuracy on plausible videos but low accuracy on implausible ones, indicating a strong tendency to default to plausible interpretations—even when physical violations are present.

With TRAVL fine-tuning, both models improve significantly on implausible videos: Video-ChatGPT increases from 0.0\% to 12.0\%, and LLaVA-NeXT from 3.3\% to 52.7\%. However, this comes at the cost of reduced accuracy on plausible videos. This suggests increased sensitivity to physical inconsistencies, but also a higher rate of false positives on real videos. However, TRAVL still performs better on both plausible and implausible videos compared to the same backbone trained via SFT. This shift in behavior may be partly due to the fine-tuning data, which is skewed toward implausible examples and contains a more limited set of real, plausible scenarios. Expanding the range of plausible examples during training may help to better calibrate model confidence across both types of videos.

\section{Fine-Tuning Dataset}
\label{supp:ftdataset}

Our fine-tuning dataset comprises \textbf{3,482 unique videos} and \textbf{19,708 question-answer (QA) pairs}. The dataset integrates four sources: (1) the Video-ChatGPT training set, (2) the Physics-IQ benchmark along with newly generated implausible variants, (3) 535 clips from the \emph{Physics} category of Impossible Videos, and (4) 92 additional implausible clips generated with Pika 1.5. Together, these sources provide a broad balance of real and generated content, and expose models to diverse motion contexts and implausibility types. 

\paragraph{Video-ChatGPT Subset.}  
We include 1,763 videos from the original Video-ChatGPT training set, filtering to those shorter than 800 frames so that motion trajectories remain temporally coherent. Each video is captioned with GPT-4o, and we regenerate QAs to provide richer detail. In addition to general video-understanding queries, we introduce plausibility-oriented questions so that both plausible and implausible clips are associated with comparable QA types. This balancing prevents models from learning shortcut correlations between question style and video category.

\paragraph{Physics-IQ Scenarios.}  
We take 66 base scenarios from Physics-IQ~\cite{physics-iq}, each recorded from three viewpoints (198 total plausible videos). These scenarios illustrate core physics principles in short real-world clips. To expand this set, we generate 894 implausible variants using image-to-video models (Pika, Runway, Sora, Kling, and Luma), conditioned on the first frame and a caption of the original scenario. Human annotators review all generations, discarding around 70 plausible or static cases, and label the retained clips by violation type (e.g., floating, teleportation). Annotators also provide concise temporal feedback (e.g., “duck disappears midair”), which GPT-4o incorporates into detailed captions and 3–6 QA pairs per video.

\paragraph{Impossible Videos Scenarios.}  
As shown in Table~\ref{tab:blindtest}, the multiple-choice questions in Impossible Videos~\cite{bai2025impossible} can be solved by LLMs using linguistic biases alone, making it unsuitable for evaluation. Instead, we repurpose 535 videos from the physics category as training data. Each video is captioned with GPT-4o, which is given access to the correct physical violation. These captions are then passed through the same QA-generation pipeline, producing 3–6 QAs per video.

\paragraph{Additional Implausible Videos.}  
Finally, we generate 92 diverse implausible clips using Pika 1.5. Captions are sampled from GPT-4o to cover a broad range of everyday scenarios. Each clip is manually inspected to ensure the presence of a clear implausibility, captioned accordingly, and passed through the same QA pipeline. This set complements Physics-IQ’s object-limited scenarios with more generic violations.

\subsection{QA Generation Prompts}
\label{supp:qaprompt}

To create fine-grained temporal and physical reasoning QA pairs, we used GPT-4o with structured prompts. Each prompt takes as input (1) the \texttt{scenario} name and (2) a manually verified caption describing the video. We design separate instructions for plausible and implausible videos to avoid leakage of implausibility cues in the questions.

\paragraph{Implausible Videos.}
For videos containing physically unrealistic events, the prompt explicitly instructs GPT-4o that the clip is implausible. The generated \emph{answers} must clearly explain why, but the \emph{questions} remain neutral. This prevents models from exploiting phrasing such as “what makes this implausible?” and ensures that implausibility is only reflected in the answers. Below is the exact prompt:

\begin{figure*}[t]
\begin{lstlisting}[language=Python, caption={Training Q/A generation prompt}, label={lst:qagen}]
prompt_messages = [
    {
        "role": "user",
        "content": f"""
    You are an expert in video-language reasoning. Your task is to generate
    3 to 6 question–answer (Q/A) pairs for the given video scenario and caption.

    All videos in this batch are implausible — they contain physically
    unrealistic events. The answers must explicitly state this and
    explain why the scene is implausible, based only on the caption.

    Questions should focus on:
    - General video understanding (overall events, including what appears implausible)
    - Physical realism (phrased neutrally, e.g., “Do the events appear realistic or implausible?”)
    - Physical behavior (object interactions, motion, deformations)
    - Temporal reasoning (what happens first, next, last)

    Instructions:
    - Generate 3 to 6 Q/A pairs per scenario. Never fewer, never more.
    - Include at least one neutral question on physical realism.
    - DO NOT ask “What makes the video implausible?” or similar.
      Implausibility should only appear in the answers.
    - Questions must sound natural and varied.
    - Answers must be detailed, grounded only in the caption, and
      list all reasons for implausibility.

    Output Format:
    Q1: <question 1>
    A1: <answer 1>
    Q2: <question 2>
    A2: <answer 2>
    ...

    Video Scenario:
    {scenario}

    Video Description:
    {caption}
    """
    }
]
\end{lstlisting}
\end{figure*}

\paragraph{Plausible Videos.}
For real videos, the prompt is nearly identical, except that it specifies the clips are physically \emph{realistic}. In this case, the answers must highlight why the events follow physical principles, again without the questions giving away plausibility.

\section{ImplausiBench Construction}
\label{supp:implausibench}

To construct ImplausiBench, we selected 150 real-world videos spanning a diverse range of everyday scenarios, including food preparation, vehicles, animals, nature, and household activities. We first used GPT-4o to generate captions for each real video and manually verified their correctness to ensure high-quality textual descriptions. Next, we created implausible counterparts for each video by prompting state-of-the-art image-to-video models using the first frame of the real video and guiding them to generate physically unrealistic continuations.

To evaluate model understanding of physical plausibility, we designed multiple-choice questions for each video pair. These questions were constructed with the explicit goal of minimizing blind-test accuracy of language models (LLMs). This involved manual crafting of challenging distractors and iterative refinement to prevent models from relying on linguistic shortcuts alone.

This makes ImplausiBench a particularly challenging benchmark: for example, Gemini 2.5-pro, the best-performing model in our experiments, only achieved 41\% on implausible videos. We envision this benchmark as a valuable progress indicator for future models aspiring to reason about physical realism in videos.

\begin{figure*}[t]
\centering
\includegraphics[width=\textwidth]{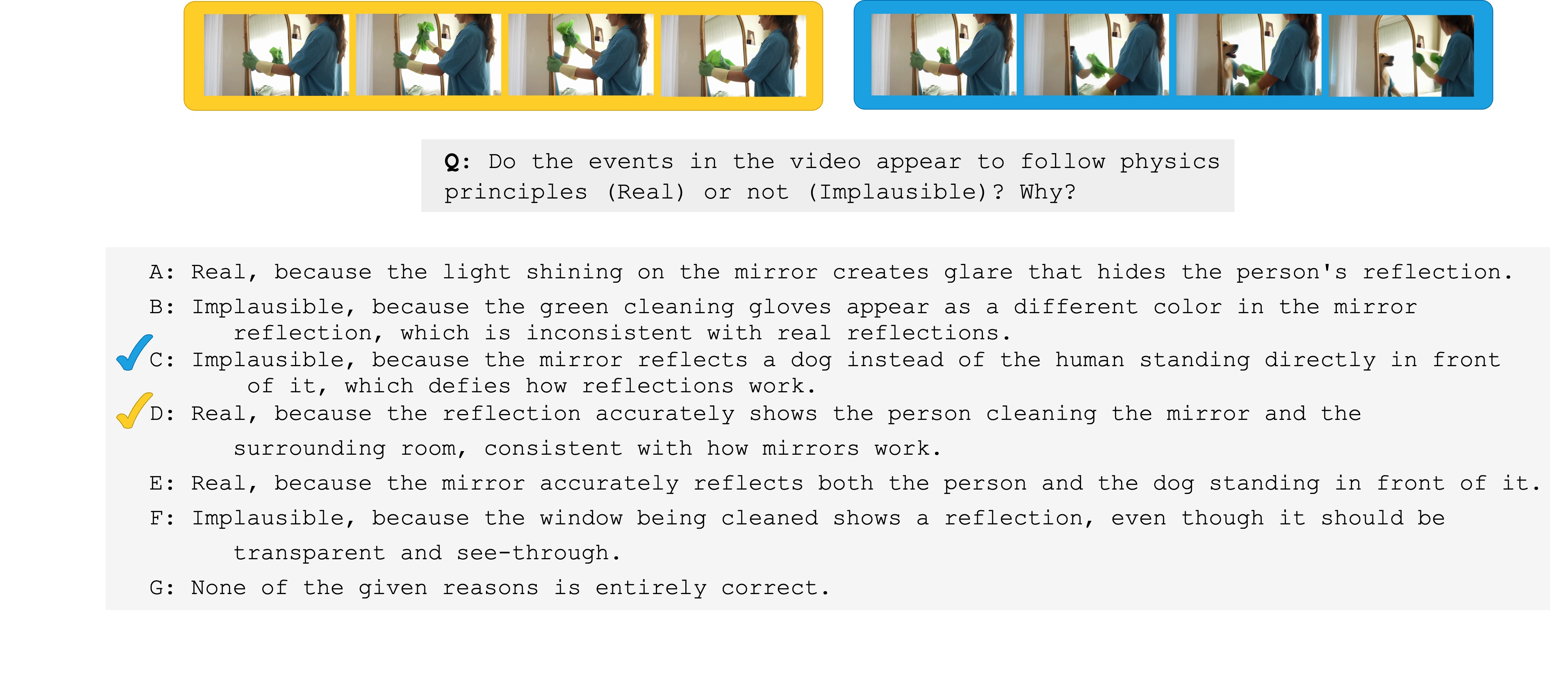}\vspace{4pt}
\includegraphics[width=\textwidth]{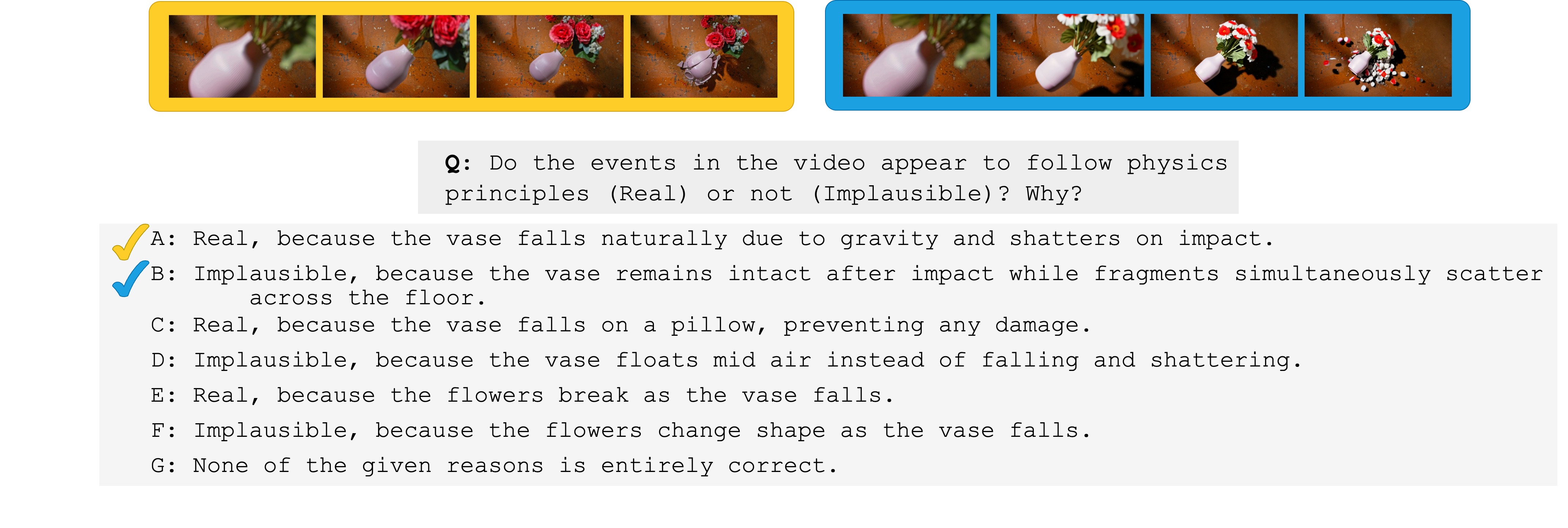}\vspace{4pt}
\includegraphics[width=\textwidth]{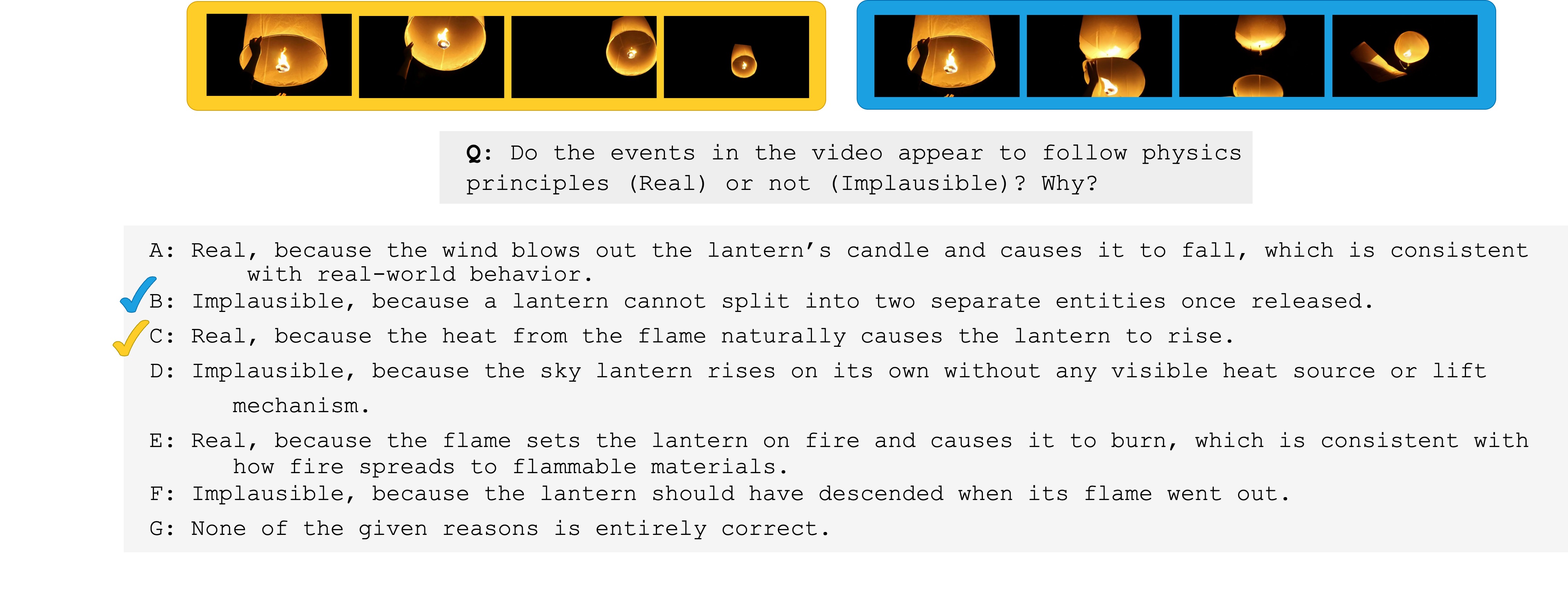}\vspace{6pt}
\caption{\textbf{Qualitative examples from ImplausiBench.}
Each row shows a real video (left) and its implausible counterpart (right). 
Pairs share a seven-option MCQ designed to prevent language-only shortcuts.}
\label{fig:implausibench-samples}
\end{figure*}

Some qualitative examples are shown in Figure~\ref{fig:implausibench-samples}. To view them in video format, please view supplementary file \texttt{result\_viewer.html}.

\section{TRAVL Details and Model Specifications}
\label{supp:travl}

\paragraph{Video-ChatGPT Integration.} 
TRAVL is inserted between the frozen CLIP encoder and the language adapter. We apply spatial self-attention within each frame (256 tokens) and trajectory-aware temporal attention across tracked patches (100 tokens). The resulting features are aggregated and projected via a 2-layer MLP to form the final 356-token sequence, which is passed to the language model.

\paragraph{LLaVA-NeXT Integration.} 
TRAVL receives 64$\times$729 SigLIP patch tokens. We apply intra-frame spatial attention for each set of 729 tokens, followed by inter-frame temporal attention over patch-aligned trajectories using sparse flow masks. To manage memory, we chunk temporal attention into overlapping windows of 4--16 frames. Features are then projected and passed to the frozen LLaVA adapter. Given that implausible actions happen suddenly (levitation, multiplication, vanishing, etc.), 16 frames is enough to detect such events. However, it is worth noting that other types of reasoning such as longer video understanding could be hurt by this shorter attention chunking.

\paragraph{Trajectory-Guided Sparse Attention Masking.}
To enable temporal reasoning over object motion, we construct a sparse attention mask using CoTracker to track the center of each spatial patch across time. Each video frame is divided into a grid of patches that matches the resolution of the vision encoder (e.g., $27 \times 27$ for LLaVA-NeXT). To account for newly appearing objects or major scene changes, we reinitialize a set of track points at the center of each patch every $k$ frames. Each tracked point is assigned a patch ID at every visible frame, and we construct a binary attention mask that connects patch pairs sharing a common trajectory. This sparse mask is then used to constrain temporal self-attention, enabling the model to focus on motion-consistent features while significantly reducing computational cost.

\paragraph{Trajectory Mask Calculation.} Pseudocode \ref{lst:maskgen} describes how we compute the patch-based trajectory mask.

\vspace{0.5em}
\begin{figure*}
\begin{lstlisting}[language=Python, caption={Mask generation code}, label={lst:maskgen}]
def generate_attention_mask(video, cotracker, k=10):
    # video: [T, 3, 384, 384]; reinit tracking every k frames
    T, _, H, W = video.shape
    # grid size 729 patches
    G = 27  
    P = G * G

    queries, q_times = [], []
    for t in range(0, T, k):
        for i in range(G):
            for j in range(G):
                x = (j + 0.5) * (W / G)
                y = (i + 0.5) * (H / G)
                queries.append([t, x, y])
                q_times.append(t)

    tracks, vis = cotracker(video[None], queries=queries, t_valid=q_times)

    patch_ids = ((tracks[0, ..., 1] // (H // G)).long() * G +
                 (tracks[0, ..., 0] // (W // G)).long())

    mask = torch.zeros((T * P, T * P), dtype=torch.bool)
    for n in range(len(queries)):
        q = q_times[n]
        p0 = patch_ids[q, n]
        for t in range(T):
            if vis[0, t, n] > 0.5:
                pt = patch_ids[t, n]
                mask[q * P + p0, t * P + pt] = True
        mask[q * P + p0, q * P + p0] = True

    mask.fill_diagonal_(True)
    return mask
\end{lstlisting}
\end{figure*}

\paragraph{Attention Mechanism.} Pseudocode \ref{lst:attn} describes the masked attention mechanism using the sparse trajectory mask.

\begin{figure*}[t]
\centering
\begin{minipage}{.98\textwidth}
\begin{lstlisting}[language=Python, caption={Attention module code}, label={lst:attn}]
def apply_travl_attention(patch_tokens, flow_mask):
    # patch_tokens: [B, T, P, D] where T=frames, P=patches, D=dim
    # mask: [B, T*P, T*P] binary mask 

    spatial_out = []
    for t in range(T):
        frame_tokens = patch_tokens[:, t]  # shape [B, P, D]
        frame_attn = self_attend(frame_tokens)  # spatial attention
        spatial_out.append(frame_attn)
    spatial_out = torch.stack(spatial_out, dim=1)  # shape [B, T, P, D]

    flat_tokens = spatial_out.view(B, T*P, D)
    attended = masked_temporal_attention(flat_tokens, mask)  # [B, T*P, D]
    return attended.view(B, T, P, D)
\end{lstlisting}
\end{minipage}
\end{figure*}

\paragraph{Training Details.}
\begin{itemize}
  \item Optimizer: AdamW
  \item Learning rate: $1\times10^{-4}$ for attention modules, $5\times10^{-5}$ for projector
  \item Batch size: 8 for Video-ChatGPT, 2 for LLaVA-NeXT
  \item Hardware: 4x NVIDIA H200 GPUs
  \item Epochs: 5
\end{itemize}

\section{Observations}
\label{supp:whatsnext}

In this section, we share key observations made during the development and experimentation of TRAVL. Our aim is to highlight practical insights and challenges that arose while adapting trajectory-aware attention for video-language models (VLMs). We hope these reflections are useful to researchers working on related problems in multimodal learning, video understanding, and physical reasoning, and that they serve as a roadmap for future iterations of TRAVL. Many of the issues we encountered relate to data scale, architecture compatibility, and training efficiency, which we discuss below in detail.

\paragraph{Fine-tuning Frame Rate.}
In our current fine-tuning dataset, we retain each video’s original frame rate (FPS). A natural extension is to augment the dataset by re-encoding videos at different FPS values. This would expose the model to a greater variety of temporal resolutions and increase the number of training frames, potentially improving the robustness of temporal attention and enhancing downstream performance.

\paragraph{Impact of Token Count.}
We explored increasing the number of tokens per frame in the Video-ChatGPT $+$ TRAVL setup. The original vision-language projector in Video-ChatGPT is trained on 356 tokens, derived from spatial and temporal pooling of CLIP patch features. To increase token granularity, we experimented with reducing the pooling stride, thus preserving more patch tokens across time. However, we consistently found that these configurations underperformed compared to the original 356-token setup. We hypothesize that this degradation stems from a mismatch with the pretrained projector, which is specialized for the 356-token format. Without reinitializing or retraining the projector from scratch, deviating from this token structure appears to hinder alignment and performance.

\paragraph{Scaling the Dataset.}
Our current dataset is currently modest in its coverage of different scenarios. Future efforts should focus on expanding the dataset not only in terms of the diversity of \textit{implausibility types}, but also with more varied and complex \textit{plausible} videos. As shown in Table~\ref{tab:implausibench}, while improving overall implausibility detection, TRAVL hurts the model’s performance on plausible videos compared to a pretrained model. We attribute this effect to the limited distribution of plausible content in the training set.

A key bottleneck in scaling arises from the lack of timestamped captions in the Video-ChatGPT training data. Without temporally grounded annotations, we cannot easily repurpose long videos into shorter segments with accurate supervision. To preserve coherent motion trajectories, we limit our training set to videos under 800 frames—ensuring that sampled frames are not spaced so far apart that motion becomes ambiguous or incoherent. Addressing this constraint remains an open challenge for future work, and we believe that improved timestamp alignment or synthetic supervision could unlock much larger and more balanced fine-tuning corpora.

\paragraph{Trajectory Masks.}
We used CoTracker to generate sparse trajectory masks for TRAVL. To account for new objects entering the scene or significant scene changes, we reinitialize patch tracking every $k$ frames. While effective, this approach introduces computational overhead, especially when applied on-the-fly during training. Due to speed constraints, we limited tracking to a single point per patch (i.e., the center pixel). However, denser tracking—e.g., tracking multiple points per patch—could potentially yield richer motion cues and further enhance the model’s understanding of dynamic interactions. Exploring more efficient or precomputed trajectory pipelines is an important direction for future work.

\paragraph{Beyond Implausibility.} Our benchmark focuses on detecting violations. A future direction is to generate and evaluate physically grounded captions, affordance predictions, or causal reasoning in video.

\end{document}